# Structured Priors for Structure Learning

**V. K. Mansinghka, C. Kemp, J. B. Tenenbaum**
Dept. of Brain & Cognitive Sciences
Massachusetts Institute of Technology
Cambridge, MA 02139
{vkm, ckemp, jbt}@mit.edu

**T. L. Griffiths**
Dept. of Cognitive & Linguistic Sciences
Brown University
Providence, RI 02912
tom_griffiths@brown.edu

## Abstract

Traditional approaches to Bayes net structure learning typically assume little regularity in graph structure other than sparseness. However, in many cases, we expect more systematicity: variables in real-world systems often group into classes that predict the kinds of probabilistic dependencies they participate in. Here we capture this form of prior knowledge in a hierarchical Bayesian framework, and exploit it to enable structure learning and type discovery from small datasets. Specifically, we present a nonparametric generative model for directed acyclic graphs as a prior for Bayes net structure learning. Our model assumes that variables come in one or more classes and that the prior probability of an edge existing between two variables is a function only of their classes. We derive an MCMC algorithm for simultaneous inference of the number of classes, the class assignments of variables, and the Bayes net structure over variables. For several realistic, sparse datasets, we show that the bias towards systematicity of connections provided by our model can yield more accurate learned networks than the traditional approach of using a uniform prior, and that the classes found by our model are appropriate.

## 1 Introduction

Unsupervised discovery of structured, predictive models from sparse data is a central problem in artificial intelligence. Bayesian networks provide a useful language for describing a large class of predictive models, and much work in unsupervised learning has focused on discovering their structure from data. Most approaches to Bayes net structure learning assume generic priors on graph structure, sometimes encoding a sparseness bias but otherwise expecting no regularities in the learned graphs. However, in many cases, we expect more systematicity: variables in real-world systems often play characteristic roles, and can be usefully grouped into classes that predict the kinds of probabilistic dependencies they participate in. This systematicity provides important constraints on many aspects of learning and inference.

Consider the domain of medical learning and reasoning. Knowledge engineers in this area have historically imposed strong structural constraints; the QMR-DT network [Shwe et al., 1991], for example, segregates nodes into *diseases* and *symptoms*, and only permits edges from the former to the latter. Recent attempts at medical knowledge engineering have continued in this tradition; for example, Kraaijeveld et al. [2005] explicitly advocate organizing Bayes nets for diagnosis into three layers, with influence flowing only from *context* through *fault* to *influence* nodes. Similar divisions into classes pervade the literature on probabilistic graphical models for biological interactions; [Pe'er et al., 2006], for example, learn gene networks by discovering a salient example, structuring the problem of learning gene networks around the discovery of a small set of *regulators* that are responsible for controlling the activation of all other genes and may also influence each other.

Knowledge about relationships between classes provide inductive constraints that allow Bayesian networks to be learned from much less data than would otherwise be possible. For example, consider a structure learner faced with a database of medical facts, including a list of patients for which a series of otherwise undifferentiated conditions have been provided. If the learner knew that, say, the first ten conditions were *diseases* and the rest *symptoms*, with influence only possible from diseases to symptoms, the learner would need to consider a dramatically reduced hypothesis space of structures. Various forms of less specific knowledge could also be quite useful, such as knowing that some variables played a similar causal role, even without the precise knowledge of what that role entailed. Causal roles also support transfer: a medical expert system knowledgeable about lung conditions faced with data about liver conditions defined over entirely new variables would not be able to transfer any specific probabilistic dependencies but could potentially transfer abstract structural patterns (e.g., that diseases cause symptoms). Finally, abstract structural knowledge is often interesting in its own right, aside from any inductive bias it contributes. For example, a biologist might be interested to learn that certain genes are regulators, over and above learning about specific predictive or causal links between particular genes.

We present a hierarchical Bayesian approach to structure learning that captures this sort of abstract structural knowledge using nonparametric block-structured priors over Bayes net graph structures. Given observations over a set of variables, we simultaneously infer both a specific Bayes net over those variables and the abstract structural features of these dependencies. We assume that variables come in one or more classes and that the prior probability of an edge existing between two variables is a function only of their classes. In general, we do not assume knowledge of the number of classes, the class assignments of variables, or the probabilities of edges existing between any given classes; these aspects of the prior must be inferred or estimated at the same time that we learn the structure and parameters of the Bayes net. We derive an approach for simultaneous inference of all these features from data: all real-valued parameters are integrated out analytically, and MCMC is used to infer the number of classes, the class assignments of variables, and the Bayes net structure over those variables. We show that the bias towards systematicity of connections provided by our model can yield more accurate recovery of network structures than a uniform prior approach, when the underlying structure exhibits some systematicity. We also demonstrate that on randomly generated sparse DAGs, the nonparametric form of our inductive bias can protect us from incurring a substantial penalty due to prior mismatch.

A number of previous approaches to learning and inference have exploited structural constraints at more abstract levels than a specific Bayesian network. These approaches typically show a tradeoff between the representational expressiveness and flexibility of the abstract knowledge, and the ease with which that knowledge can be learned. At one end of this tradeoff, approaches like [Segal et al., 2005] assume that the Bayes net to be learned must respect very strict constraints — in this case, that variables can be divided into modules, where all variables in a module share the same parents and the same conditional probability table (CPT). This method yields a powerful inductive bias for learning both network structure and module assignments from data, especially appropriate for the study of gene regulatory networks, but it is not as appropriate for modeling many other domains with less regular structure. The QMR-DT [Shwe et al., 1991] or Hepar II [Kraaijeveld et al., 2005] networks, for example, contain highly regular but largely nonmodular structure, and we would like to be able to discover both this sort of network and a good characterization of this sort of regularity. At the other end of the flexibility-learnability tradeoff, frameworks such as probabilistic relational models (PRMs) [Friedman et al., 1999] provide a far more expressive language for abstract relational knowledge that can constrain the space of Bayesian networks appropriate for a given domain, but it is a significant challenge to learn the PRM itself — including the class structure — from raw, undifferentiated event data. Our work aims at a valuable intermediate point on the flexibility-learnability spectrum. We consider hypotheses about abstract structure that are less expressive than those of PRMs, but which are simultaneously learnable from raw data along with specific networks of probabilistic dependencies. Our nonparametric models also yield a stronger inductive bias for structure learning than a uniform prior but a weaker and more flexible one than module networks.

## 2 Block-Structured Priors

The intuition that nodes in a Bayes net have predictive classes can be formalized in several ways. In this paper, we focus on two structure priors obtained from similar formalizations: the *ordered blockmodel* and the *blockmodel*. The joint distributions we work in (including the uniform baseline) can be described by a meta-model consisting of three major pieces (see Figure 1): a prior over the structure of Bayes nets, a prior on the parameterizations of Bayes nets given their structure, and the data likelihood induced by that parameterization. For simplicity, we work with discrete-state Bayesian networks with known domains, and use the standard conjugate Dirichlet-Multinomial model for their conditional probability tables [Cooper and Herskovits, 1992]. Note that the variable $G$ in Figure 1 refers to a graph structure, with $B$ its elaboration into a full Bayesian network with conditional probability tables. Inference in the model of Figure 1 and its specializations characterizes the Bayes net structure learning problem.

We are interested in discovering node classes that explain patterns of incoming and outgoing edges, as a step towards representation and learning of causal roles. Our starting point is the *infinite blockmodel* of [Kemp et al., 2004], a nonparametric generative model for directed graphs, which we modify in two ways to only produce acyclic graphs. We first describe the generative process for the ordered blockmodel:

1. Generate a class-assignment vector $\vec{z}$ containing a partition of the $N$ nodes of the graph via the Chinese restaurant process or CRP [Pitman, 2002] with hyperparameter $\alpha$. The CRP represents a partition in terms of a restaurant with a countably infinite number of tables, where each table corresponds to a group in the partition (see Figure 1b) and the seating assignment of person $i$ is the class of the $i$th object. People are seated at each existing table $k$ (from 1 up to $K^+$) with probability proportional to $m_k$, the number of previous occupants of the table. New tables are created with probability proportional to the hyperparameter $\alpha$. As the CRP is exchangeable, the distribution on partitions (represented by $\vec{z}$) that it induces is invariant to the order of entry of people into the restaurant. Specifically, we have:

$$P(\vec{z}|\alpha) = \alpha^{K^+} \frac{\Gamma(\alpha)}{\Gamma(N+\alpha)} \prod_{k=1}^{K^+} (m_k - 1)! \qquad (1)$$

2. Generate an ordering $\vec{o}$ of the $K^+$ classes in the partition $\vec{z}$ uniformly at random:

$$P(\vec{o}|\vec{z}) = \frac{1}{K^+!} \qquad (2)$$

   Then $o_a$ contains the order of class $a$.

3. Generate a square graph template matrix $\eta$, of size equal to the number of classes in the generated partition, where $\eta_{o_a,o_b}$ represents the probability of an edge between a node of class $a$ and a node of class $b$. To ensure acyclicity, fix all entries to be 0 except those strictly above the diagonal; these possibly nonzero entries $\eta_{o_a,o_b}$ with $o_b > o_a$ are drawn from a $Beta(\vec{\beta})$ distribution (with $B(\cdot,\cdot)$ its normalizing constant):

$$P(\eta_{o_a,o_b}|\vec{z}) = \frac{1}{B(\beta_1,\beta_2)} \eta_{o_a,o_b}^{\beta_1-1}(1-\eta_{o_a,o_b})^{\beta_2-1} \quad (3)$$

4. Generate a graph $G$ by drawing each edge $G_{i,j}$ from a Bernoulli distribution with parameter $\eta_{o_{z_i},o_{z_j}}$.

$$P(G|\vec{z},\vec{o},\eta) = \prod_{i=1}^{N}\prod_{j=1}^{N} \eta_{o_{z_i},o_{z_j}}^{G_{i,j}} (1-\eta_{o_{z_i},o_{z_j}})^{1-G_{i,j}}$$

Note that all graphs are possible under this process, though the ones lacking salient block structure typically require more classes. Also note that the hyperparameters in this process are intuitive and interpretable. $\alpha$ controls the prior on partitions of nodes into classes; smaller values favor fewer groups *a priori*. $\vec{\beta}$ controls the $\eta$ matrix, with values such as (0.5, 0.5) favoring clean blocks (either all edges absent or all present) and increases in $\beta_1$ or $\beta_2$ yielding biases towards sparseness or denseness respectively. The ordering $\vec{o}$ encodes the "causal depth" of nodes based on their classes. Because we believe specific, directed dependencies are important outputs of structure discovery for causal systems, we do not constrain our priors to assign equal mass to members of a particular Markov equivalence class.

The blockmodel prior does not include the ordering $\vec{o}$. Graph generation is essentially as above, with $\eta_{a,b}$ being the probability of an edge between a node of class $a$ and a node of class $b$, and all entries permitted to be nonzero. Cyclic graphs are rejected afterwards by renormalization. More formally, for the blockmodel version, let $\Delta(G)$ be 1 if $G$ is acyclic and 0 otherwise. Then we have

$$P(G|\vec{z},\eta) \propto \Delta(G) \prod_{i=1}^{N}\prod_{j=1}^{N} \eta_{z_i,z_j}^{G_{i,j}} (1-\eta_{z_i,z_j})^{1-G_{i,j}} \quad (4)$$

Both priors may be appropriate depending on the situation. If we are interested in classes that correspond to different stages (possibly unrolled in time) of a causal process, the ordered blockmodel may be more appropriate. If we expect nodes of a given class to include some connections with others of the same class, as in the study of gene regulatory networks where some regulators influence others, then the blockmodel should perform better. Additionally, both models can generate any directed acyclic graph. The blockmodel can do this either by assigning all nodes to the same class, learning the sparsity of the graph, or by assigning all nodes to different classes (which may be useful for dense, nearly cyclic graphs). The ordered blockmodel has the option of assigning nodes to different classes corresponding to layers in the topological ordering of an arbitrary DAG.

Our data likelihood is the standard marginal likelihood for complete discrete observations, assuming the CPTs (represented by the parameters $B$) have been integrated out; $\gamma$ plays the role of the pseudo-counts setting the degree of expected determinism of the CPTs [Murphy, 2001]. Overall, then, our model has three free parameters (in the symmetric $\beta$ case) — $\alpha$, $\beta$ and $\gamma$ — which each give us the opportunity to encode weak prior knowledge at different levels of abstraction.

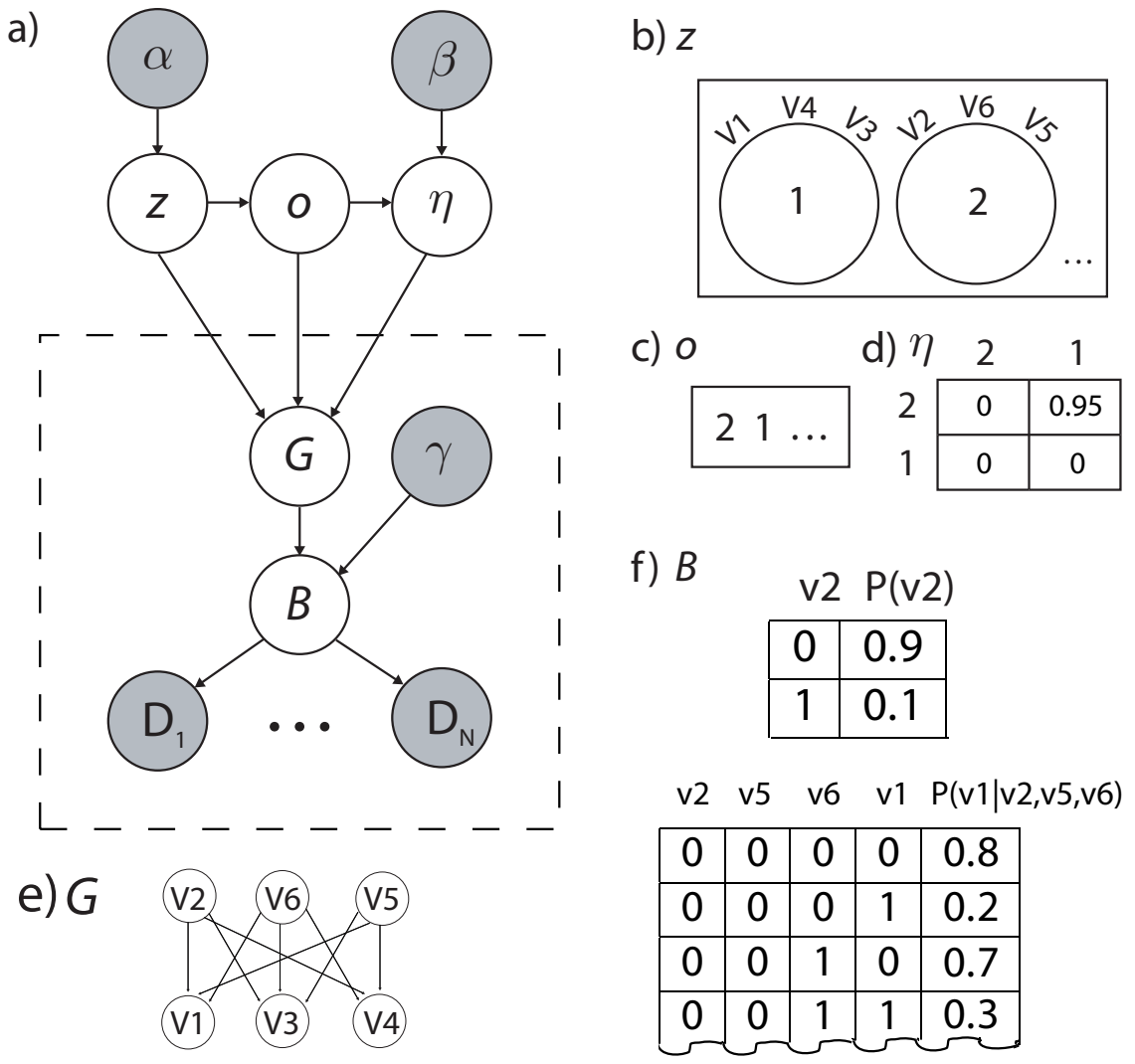

Figure 1: (a) Graphical meta-model for the *ordered blockmodel*. The dashed line indicates the components of the meta-model corresponding to traditional Bayes net structure learning with a uniform prior. (b)-(d) Latent variables representing abstract structural knowledge.

This nonparametric, hierarchical approach is attractive for several reasons. The CRP gracefully handles a spectrum of class granularities, generally preferring to produce clumps but permitting each object to be in its own class if necessary. $\vec{z}$, $\vec{o}$ and $\eta$ provide convenient locations for the insertion of strong or weak prior knowledge, if we have it, ranging from complete template knowledge to only the expectation that some reasonable template can be found. They also represent additional outputs of learning, which are of interest in both cognitive applications (e.g. the concept *disease* is at least as much about the patterns in the causal relationships present across many particular diseases as it is about those relationships in particular) and in scientific data analysis (e.g. the notion that regulators exist is useful even if one isn't entirely sure about which genes are regulated by which other ones).

Throughout this paper, we will compare our approach with Bayes net structure learning given a conventional, uniform prior on DAGs. Formally, this corresponds to the subcomponent of the meta-model in Figure 1 indicated by the dashed box, and a prior on graph structures $G$ given by Equation 4 with a single class and $\eta$ (now a scalar) fixed to 0.5.

## 3 Inference

Having framed our approach to structure learning as inference in the graphical model of Figure 1, we must provide an effective procedure for performing it. Note that when determining the contribution of a given structure to the predictive distribution, we will explicitly represent the posterior on parameters for that structure, again using conjugacy. To further reduce the size of our hypothesis space for the experiments in this paper, we also integrate out the $\eta$ matrix. Let $n^+_{a,b}$ be the number of edges between nodes of class $a$ and nodes of class $b$, $n^-_{a,b}$ the corresponding number of missing edges, with $K^+$ the number of classes as before. Then we have:

$$p(G|\vec{z}) = \prod_{a=1}^{K^+} \prod_{b=1}^{K^+} \frac{B(\beta_1 + n^+_{a,b}, \beta_2 + n^-_{a,b})}{B(\beta_1, \beta_2)} \tag{5}$$

Note that in the case of the blockmodel, where $\eta$ is not itself constrained to generate acyclic graphs, this equality is instead only a proportionality, due to renormalization of the prior over acyclic graphs.

Inference, in general, will entail finding posterior beliefs about $\vec{z}$, $\vec{o}$ and $G$. We organize our inference around a Markov chain Monte Carlo procedure, consisting of both Gibbs and Metropolis-Hastings (MH) kernels, to retain theoretical correctness guarantees in the limit of long simulations. Of course, it is straightforward to anneal the Markov chain and to periodically restart if we are only interested in the MAP value or a set of high-scoring states for selective model averaging [Madigan et al., 1996].

Our overall MCMC process decouples into moves on graphs and, if relevant, moves on the latent states of the prior (a partition of the nodes, and in the ordered case, an ordering of the groups in the partition). Our graph moves are simple: we Gibbs sample each potential edge, conditioned on the rest of the hidden state. This involves scoring only two states — the current graph and the graph with a given edge toggled — under the joint, renormalizing, and sampling, and is therefore quite computationally cheap. As all our priors place 0 probability mass on cyclic graphs, cyclic graphs are never accepted. Furthermore, much of the likelihood can be cached, since only nodes whose parent set changed will contribute a different term to the likelihood. We found these Gibbs moves to be more effective than the classic neighborhood based graph moves discussed in [Giudici and Castelo, 2003], though in general we expect combinations of our moves, MH-moves that randomly propose reversing edges, and neighborhood-based moves to be most effective.

Conditionally sampling the latent states that describe the block structure is also straightforward. In the unordered case, we simply use the standard Gibbs sampler for the Chinese restaurant process, fixing the class assignments of all nodes but one — call it $i$ — and running the free node through the CRP [Rasmussen, 2000]. Specifically, as the process is exchangeable, we can Gibbs sample a given entry by taking $i$ to be the last person to enter the restaurant, and sampling its class assignment $z_i$ from the normalized product of the CRP conditionals (below) and Equation 5:

$$P(z_i = k|\alpha, \vec{z}^{\,-i}) = \begin{cases} \frac{m_k}{i-1+\alpha} & \text{if } m_k > 0 \\ \frac{\alpha}{i-1+\alpha} & k \text{ is a new class} \end{cases} \quad (6)$$

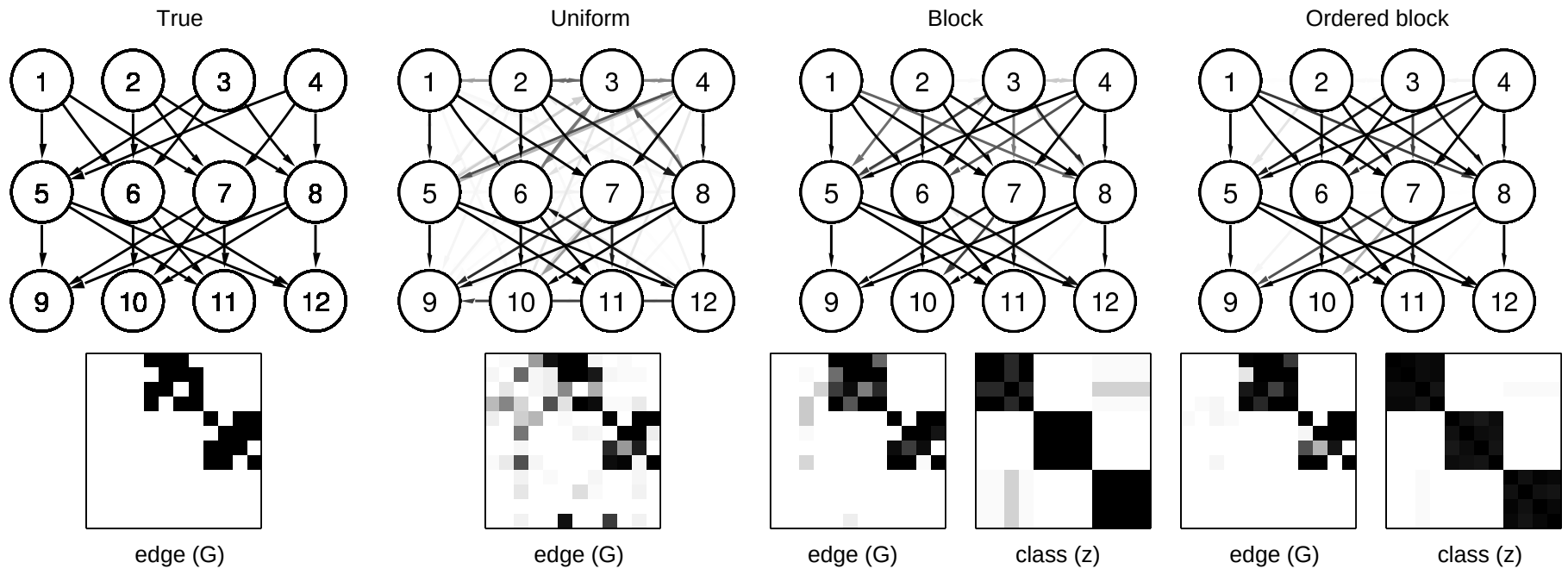

Figure 2: Results for three models on a 12-node layered topology with 75 data samples provided for training. The upper row displays the marginal probabilities of each edge in the graph under each model, obtained via selective model averaging over 100 samples. The bottom row displays the same edge marginals as an adjacency matrix, along with the marginal probabilities that $z_i = z_j$, indicating inferences about the class structure.

In the ordered case, we must be more careful. There, we fix the relative ordering of the classes of all objects except $i$, whose $z_i$ we will be resampling. When considering the creation of a new class, we consider insertions of it into $\vec{o}$ at all possible free spots in the relative order. This leaves us with an exhaustive, mutually exclusive set of possibilities for $\vec{z}$ and $\vec{o}$ which we can score under the joint distribution and renormalize to get a Gibbs sampler.

For very large problems, obtaining even approximate posteriors is beyond our current computational capabilities, but search techniques like those in [Teyssier and Koller, 2005] would no doubt be very useful for approximate MAP estimation. Furthermore, additional temperature and parallelization schemes, as well as more sophisticated moves, including splitting and merging classes, could all be used to improve mixing.

## 4 Evaluation

We evaluate our approach in three ways, comparing throughout to the *uniform* model, a baseline that uses a uniform prior over graph structures. First we consider simple synthetic examples with and without strong block structure. Second, we explore networks with topologies and parameterizations inspired by networks of real-world interest, including the QMR-DT network and a gene regulatory network. Finally, we report model performance on data sampled from HEPAR II [Onisko et al., 2001], an engineered network that captures knowledge about liver disorders.

Because our interest is in using structured priors to learn predictive structure from very small sample sizes, we usually cannot hope to identify any structural features definitively. We thus evaluate learning performance in a Bayesian setting, looking at the marginal posterior probabilities of edges and class assignments. We have focused our initial studies on fairly small networks (between 10 and 40 variables) where approximate Bayesian inferences about network structure can be done quickly, but scaling up to larger networks is an important goal for future work.

In all cases, we used our MCMC scheme to explore the space of graphs, classes, and orders, and report results based on an approximate posterior constructed from the relative scores of the 100 best models found. The pool of models we chose from was typically constructed by searching 10 times for 2000 iterations each; we found this was generally sufficient to find a state that at least matched the score of the ground truth structure.

We sampled several training and test sets from each structure; here, we report representative results. Hyperparameters were set to the same values throughout: $\alpha = 0.5, \beta = 1.0$ and $\gamma = 0.5$. For real-world applications we might adjust these parameters to capture prior knowledge that the true graph is likely to be sparse, and that the number of underlying classes is either large or small, but fixing the hyperparameters allows a fairer comparison with the uniform model.

Compared to the uniform model, we expect that the block models will perform well when the true graph is block-structured. Figure 2 shows results given 75 samples from the three-layered structure on the left. CPTs for the network were sampled from a symmetric Dirichlet distribution with hyperparameter 0.5. The true graph is strongly block structured, and even though the number of samples is small, both block models discover the three classes, and make accurate predictions about the edges that appear in the true graph. The inferences made by the uniform model are less accurate: in particular, it is relatively confident about the existence of some edges that violate the feed-forward structure of the true network. No prior will be appropriate for all datasets, and we expect that the uniform model will beat the block models in some cases when the true graph is not block-structured. Ideally, however, a learning algorithm should be flexible enough to cope with many kinds of data, and we might hope that the performance of the block models does not degrade too badly when the true graph does not match their priors. To explore this setting we generated data from graphs with sparse but non-block-structured connectivity, like the one shown in Figure 3. This graph was sampled by including each edge with probability 0.3, where we rejected all cyclic graphs and all graphs with in- or out-degrees greater than 4; CPTs were sampled from a Dirichlet distribution with hyperparameter 0.5. As expected, the block model discovers no block structure in the data, remaining confident throughout that all nodes belong to a single class. To our surprise, the block model performed better than the uniform model, making fewer mistaken predictions about the edges that appear in the true structure, and matching the true distribution more closely than the uniform model. Even though the block model found only one class, it learned the density of connections is low, winning over the uniform model (which reserves much of its probability mass for highly connected graphs) on very sparse datasets. Since the ordered block model allows no connections within classes, it cannot offer the same advantage, and Figure 3 shows that its performance is comparable to that of the uniform model.

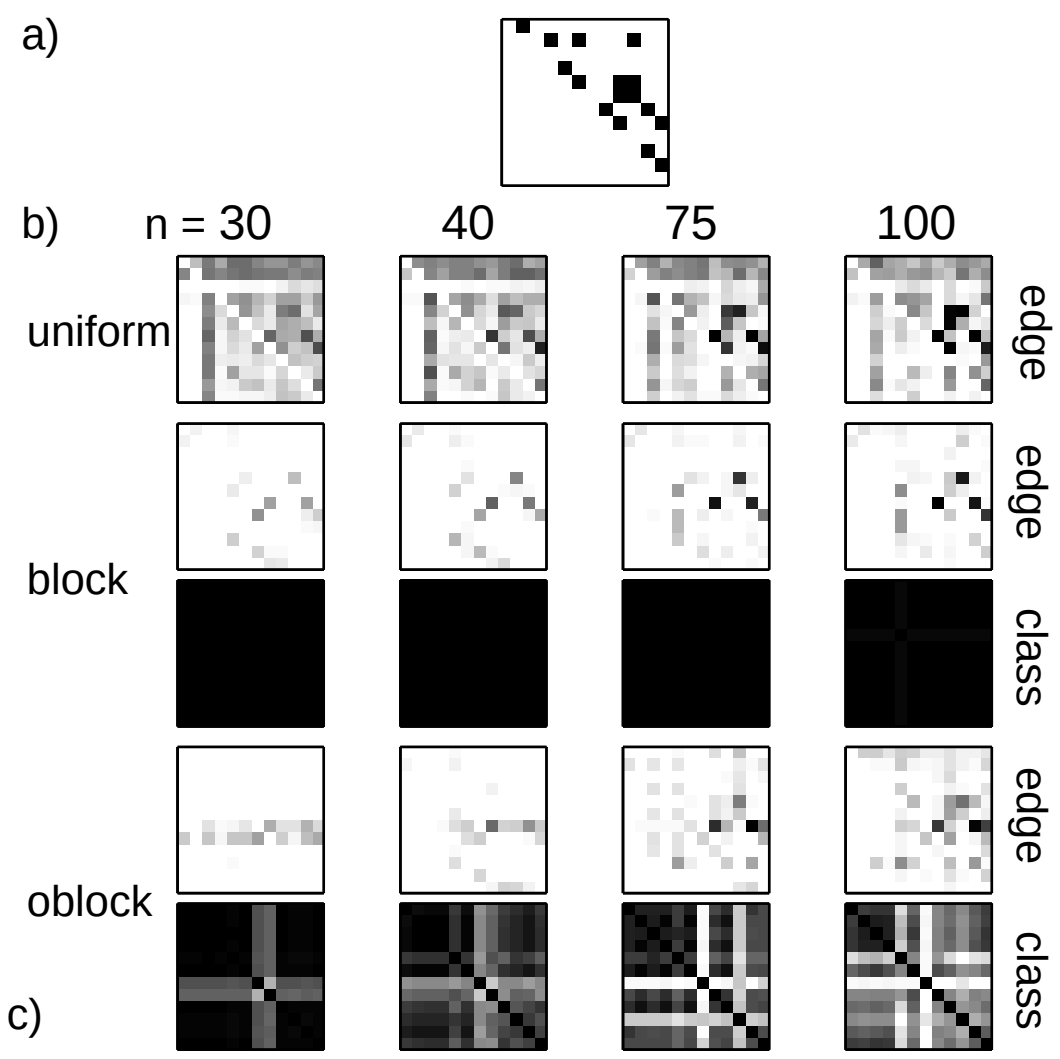

c)

| | 30 | 40 | 50 | 75 | 100 | 1000 |
|---|---|---|---|---|---|---|
| U | 0.78 | 0.60 | 0.49 | 0.44 | 0.38 | 0.03 |
| B | **0.53** | **0.37** | **0.31** | **0.33** | 0.33 | **0.03** |
| OB | 0.71 | 0.64 | 0.60 | 0.34 | **0.32** | 0.03 |

Figure 3: Learning results for data sampled from a graph without block structure. (a) adjacency matrix representing the true graph (the $(i, j)$ entry is black if there is an edge from $i$ to $j$) (b) edge and class assignment marginal probabilities for a subset of the sample sizes, as in Figure 2. (c) estimated Kullback-Leibler divergences between the posterior predictive distributions of each model and ground truth.

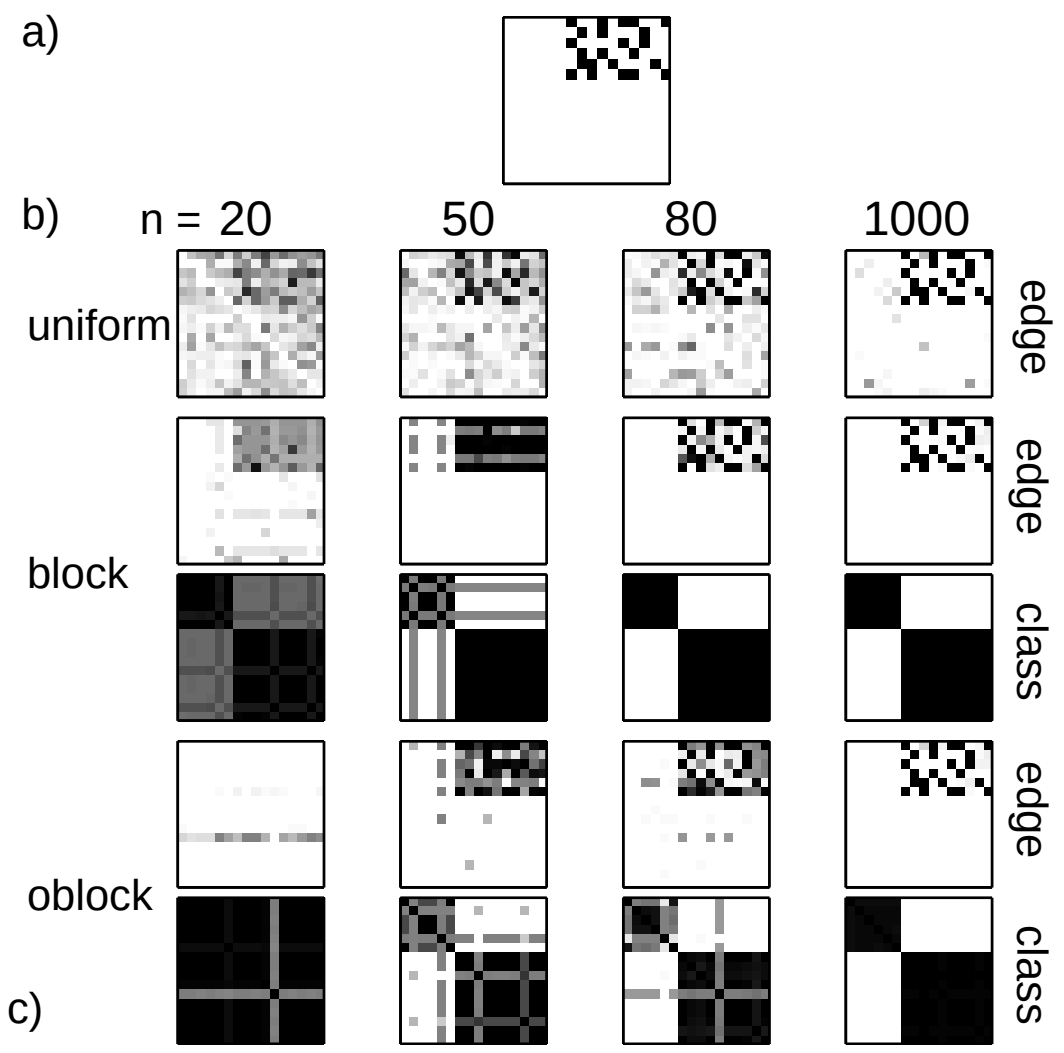

c)

| | 20 | 30 | 50 | 80 | 100 | 1000 |
|---|---|---|---|---|---|---|
| U | 1.56 | 1.34 | **0.82** | 0.50 | 0.48 | 0.05 |
| B | **1.19** | **1.12** | 1.08 | **0.28** | **0.26** | 0.04 |
| OB | 2.32 | 2.11 | 1.03 | 0.48 | 0.82 | **0.04** |

Figure 4: Learning results for data sampled from a noisy-OR QMR-like network.

Our approach to modeling abstract structure was motivated in part by common-sense medical knowledge, and networks like QMR-DT. The QMR network is proprietary, but we created a QMR-like network with the connectivity shown in Figure 4. The network has two classes, corresponding to diseases and the edges all appear between diseases and symptoms. CPTs for the network were generated using a noisy-or parameterization. The results in Figure 4 show that both block models recover the classes given a small set of examples, and it is particularly striking that the unordered model begins to make accurate predictions about class membership when only 20 examples have been provided. By the time 80 examples have been provided, the blockmodel makes accurate predictions about the edges in the true graph, and achieves a predictive distribution superior to that of the uniform model.

As [Segal et al., 2005] have shown, genetic expression data may be explained in part by the existence of underlying classes, or modules. Unlike the QMR example, links within classes should be expected: in par-ticular, we expect that genes belonging to the class *regulator* will sometimes regulate each other. To test our models on a simple setting with this structure, we generated data from the network shown in Figure 5, with CPTs sampled from a Dirichlet distribution with hyperparameter 0.5. After 1000 samples, the block model is confident of the correct class structure (with 4 regulators), and is considering one of the two correct within-regulator edges (between variables 1 and 2), although it is uncertain about the orientation of this edge. The uniform model has similar beliefs about this edge after 1000 samples, but it has equally confident incorrect beliefs about other influences on the regulators. The ordered blockmodel begins to infer class differences earlier, and is considering grouping various subsets of the regulators.

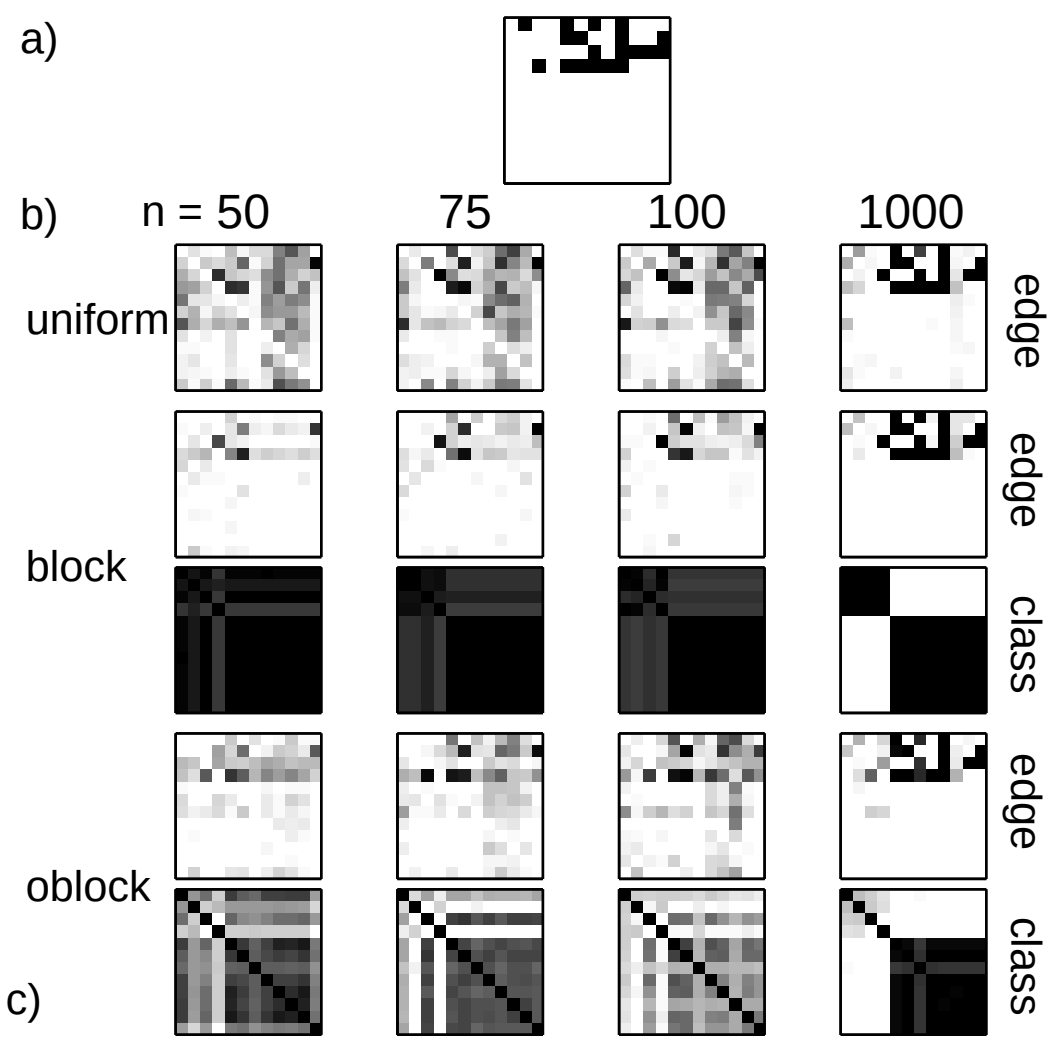

c)

| | 30 | 40 | 50 | 75 | 100 | 1000 |
|---|---|---|---|---|---|---|
| U | 1.16 | 0.79 | 0.72 | 0.68 | 0.49 | 0.04 |
| B | **0.73** | **0.64** | **0.49** | **0.42** | **0.35** | **0.03** |
| OB | 1.60 | 0.79 | 0.62 | 0.68 | 0.42 | 0.03 |

Figure 5: Learning results for data sampled from a model inspired by a genetic regulatory network.

As already suggested, knowledge about medical conditions can sometimes be organized as knowledge about interactions between three classes of variables: *risk factors*, *diseases* and *symptoms*. The HEPAR II network captures knowledge about liver diseases in a structure close to this three-part template. The structure of HEPAR II was elicited from medical experts, and the CPTs were learned from a database of medical cases. We generated training data from a subset of the full network that includes the 21 variables that appear in Figure 1 of Onisko et al. [2001] and all edges between these nodes from the true model. The network structure is shown in Figure 6. When provided with 1000 training examples, the unordered block model finds two classes, one which includes 4 of the 5 diseases, and a second which includes all the remaining variables. Note that the missing disease has a pattern of connections that is rather unusual: unlike the other 4 diseases, it has a single outgoing edge, and that edge is sent to another disease rather than a symptom. The model fails to distinguish between the risk factors and the symptoms, perhaps because the connectivity between the risk factors and the diseases is very sparse. Although neither block model recovers the three part structure described by the creators of the network, both discover sufficient structure to allow them to match the generating distribution better than the uniform model.

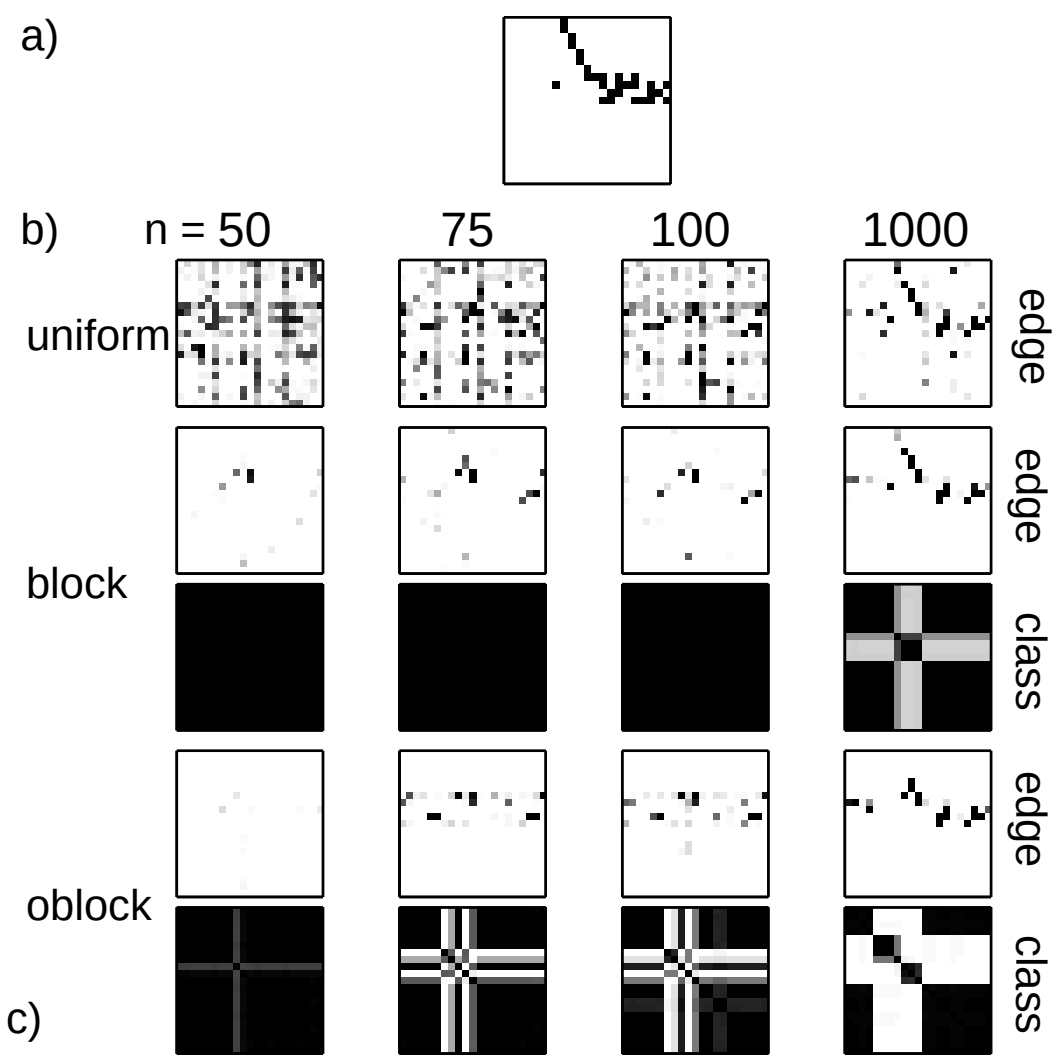

c)

| | 20 | 30 | 50 | 75 | 100 | 1000 |
|---|---|---|---|---|---|---|
| U | 2.88 | 2.21 | 1.51 | 1.01 | 0.70 | 0.07 |
| B | **2.01** | **1.62** | **1.06** | **0.58** | 0.53 | **0.06** |
| OB | 2.06 | 1.66 | 1.36 | 0.63 | **0.44** | 0.06 |

Figure 6: Learning results for data sampled from a subset of the HEPAR II network.

## 5 Discussion and Conclusions

In this paper we have explored two formalisms for representing abstract structural constraints in Bayes nets and shown how this knowledge can support structure learning from sparse data. Both formalisms are based on nonparametric, hierarchical Bayesian models that discover and characterize graph regularities in terms of node classes. We have seen how this approach succeeds when block structure is present without incurring a significant cost when it is not. We have also seen how this approach can, at the same time as it is learning Bayes net structure, recover abstract class-based patterns characterizing an aspect of the causal roles that variables play.

Our approach admits several straightforward exten-sions. First, the blockmodel prior, without modifications for acyclicity, should be directly applicable to discovering latent types in undirected graphical models and Markov models for time series, such as dynamic Bayesian networks or continuous-time Bayesian networks. Second, we expect our approach to provide additional benefits when observational data is incomplete, so that prior knowledge about likely roles becomes more significant. Such data could be incorporated via approximations to the marginal likelihood such as those from [Beal and Ghahramani, 2003].

An exploration of richer pattern languages should also be useful. For example, the notion that nodes of a certain class *can* connect to nodes of another type (but may do so only rarely, or with parameters unrelated to the types), might be more appropriate for some domains. The literature on social networks, e.g. [Wasserman and Faust, 1994], provides many examples of similar, interesting relational patterns ripe for probabilistic formalization. Closer dependence of parameterization on class would also be interesting to explore: for example, some classes might project only excitatory or inhibitory edges. Finally, we note that transfer to new systems of variables, an important function of abstract knowledge, could be implemented by adding another layer to our hierarchical model. In particular, we could flexibly share causal roles across entirely different networks by replacing our CRP with the Chinese restaurant franchise of [Teh et al., 2004].

A central feature of this work is that it attempts to negotiate a principled tradeoff between the expressiveness of the space of possible abstract patterns (with the attendant advantages of particular patterns as inductive constraints) and the possibility of learning the abstract patterns themselves. Our nonparametric approach was inspired by a striking capacity of human learning, which should also be a desideratum for any intelligent agent: the ability to learn certain kinds of "simple" or "natural" structures very quickly, while still being able to learn arbitrary — and arbitrarily complex — structures given enough data. We expect that exploring these tradeoffs in more detail will be a promising area for future research.

### Acknowledgements

We thank N. Goodman, K. Murphy and D. Koller for helpful discussions, M. Gordon for graphics assistance, and the DARPA CALO project, the James S. McDonnell Foundation Causal Learning Collaborative and NTT Communication Sciences Laboratory for financial support.